\documentclass[journal]{IEEEtran}
\usepackage{cite}

\usepackage{graphicx}

\usepackage{subfig}

\usepackage{amsmath}

\usepackage{multirow}

\ifCLASSINFOpdf
\else
\fi
\begin{document}
%
\title{Traffic Sign Timely Visual Recognizability Evaluation Based on 3D Measurable Point Clouds}
%
%
%

\author{Shanxin~Zhang,
        Cheng~Wang,~\IEEEmembership{Senior~Member,~IEEE,}
        Zhuang~Yang,~\IEEEmembership{Member,~IEEE,}
        Chenglu~Wen,~\IEEEmembership{Member,~IEEE,}
        Jonathan~Li,~\IEEEmembership{Senior~Member,~IEEE}
        and~Chenhui~Yang~
\thanks{S. Zhang is with Fujian Key Laboratory of Sensing and Computing for Smart City,
	 School of Information Science and Engineering, Xiamen University, Xiamen 361005, China,
	 and also with Xizang Key Laboratory of Optical Information Processing and Visualization Technology, 
	 Information Engineering College, Xizang Minzu University, Xianyang 712000, China (e-mail: jaysean1688@gmail.com).}
\thanks{C. Wang , Z. Yang, C. Wen, C. Yang are with Fujian Key Laboratory of Sensing and Computing for Smart City,
 	School of Information Science and Engineering, Xiamen University, Xiamen 361005, China (e-mail: cwang@xmu.edu.cn; zhuangyng@163.com;
 	clwen@xmu.edu.cn; chyang@xmu.edu.cn).}
\thanks{J. Li is with the MoE Key Laboratory of Underwater Acoustic Communication and Marine Information Technology,
	School of Information Science and Engineering, Xiamen University, Xiamen 361005, China, 
	and also with the Department of Geography and Environmental Management, Faculty of
	Environment, University of Waterloo, Waterloo, ON N2L 3G1, Canada (e-mail: junli@xmu.edu.cn).}}

%
%

\markboth{Journal of \LaTeX\ Class Files,~Vol.~14, No.~8, August~2015}%
{Shell \MakeLowercase{\textit{et al.}}: Bare Demo of IEEEtran.cls for IEEE Journals}
%



\maketitle

\begin{abstract}
The timely provision of traffic sign information to drivers is essential for the drivers to respond, to ensure safe driving, and to avoid traffic accidents in a timely manner. We proposed a timely visual recognizability quantitative evaluation method for traffic signs in large-scale transportation environments. To achieve this goal, we first address the concept of a visibility field to reflect the visible distribution of three-dimensional (3D) space and construct a traffic sign Visibility Evaluation Model (VEM) to measure the traffic sign's visibility for a given viewpoint. Then, based on the VEM, we proposed the concept of the Visual Recognizability Field (VRF) to reflect the visual recognizability distribution in 3D space and established a Visual Recognizability Evaluation Model (VREM) to measure a traffic sign's visual recognizability for a given viewpoint. Next, we proposed a Traffic Sign Timely Visual Recognizability Evaluation Model (TSTVREM) by combining VREM, the actual maximum continuous visual recognizable distance, and traffic big data to measure a traffic sign’s visual recognizability in different lanes. Finally, we presented an automatic algorithm to implement the TSTVREM model through traffic sign and road marking detection and classification, traffic sign environment point cloud segmentation, viewpoints calculation, and TSTVREM model realization. The performance of our method for traffic sign timely visual recognizability evaluation is tested on three road point clouds acquired by a mobile laser scanning system (RIEGL VMX-450) according to Road Traffic Signs and Markings (GB 5768-1999 in China) , showing that our method is feasible and efficient. 
\end{abstract}


%
\IEEEpeerreviewmaketitle

\section{Introduction}
%
%
%
%


\IEEEPARstart{T}{raffic} signs include a number of important traffic information, such as speed restrictions, driving behavior restrictions, changes ahead of road conditions and other information; the timely provision of this information to drivers increases the likelihood that the drivers will respond in a timely manner, to ensure safe driving, and to avoid traffic accidents
\cite{liu2010cognitive, kirmizioglu2012comprehensibility, ben2015effect}. Detecting and reading a roadside on-premise sign by a driver involves a complex series of sequentially occurring events, both mental and physical. They include message detection and processing, intervals of eye and/or head movement alternating between the sign and the roadway environment, and finally, active maneuvering of the vehicle (such as lane changes, deceleration, and turning into a destination) as required in response to the stimulus provided by the sign \cite{bertucci2006sign}. In these complex procedures, it is of paramount importance that traffic signs be clearly visible to the driver. However, some signs are damaged by humans or nature and some signs are occluded by other objects in the traffic environment. This may lead to a sign being of low visibility or invisible, thereby decreasing its visual recognizability and increasing the probability of a traffic accident. An efficient traffic sign timely visual recognizability evaluation method is needed to judge whether a traffic signs can be recognized in the driving process.

There are many facts that affect a traffic sign's visual recognizability for a given traffic sign in a given traffic environment. We summarize these into two aspects: objective factors and subjective factors. For objective factors, a traffic sign's size, placement, mounting height, aiming, depression angle, shape damage degree, traffic sign occlusion degree, road curve, road surface up and down, and visual continuity in the surrounding environment are the factors that affect the driver's ability to achieve visual recognition. As is known, a reasonable and legitimate traffic sign can provide the driver with good retinal imaging to help with visual recognizability. As for the subjective factors, the driver's vehicle speed, sight direction, and Viewer Reaction Time (VRT) \cite{bertucci2006sign} are the factors that affect visual recognizability. The driver’s Geometric Field Of View (GFOV) decreases progressively with increasing vehicle speed \cite{mourant2007optic}. The direction of the line of sight determines whether the traffic sign falls within the GFOV. Obviously, because of occlusion, the frequency of visual continuity being interrupted is also one of the factors that affects the traffic signs being recognized by drivers \cite{tieri2015mere}. The sign cannot be recognized by humans when the actual maximum recognizable distance is less than the Viewer Reaction Distance (VRD) \cite{bertucci2006sign}.

Of course, there are facts included by objective factors and subjective factors that also affect traffic signs’ visual recognizability and were not mentioned above. For example, the weather condition \cite{belaroussi2014impact}, light influence caused by solar elevation angle, age and sight of driver \cite{roge2004influence}, and cognitive burden of traffic density \cite{costa2014looking}, among others. We unify these as other factors. An interface is left in the TSTVREM model regarding other factors’ influence to research them in the future.

\begin{figure}[!t]
	\centering
	\subfloat[Object occlusion]{\includegraphics[width=1in,height=1.4in]{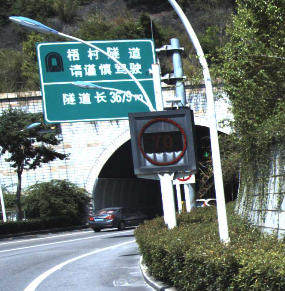}%
	\label{fig_PicObjectOcclustion}}
	\hfil
	\subfloat[Object occlusion]{\includegraphics[width=1in,height=1.4in]{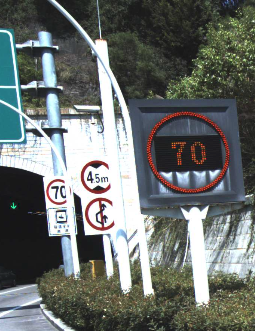}%
	\label{fig_PicObjectOcclustion2}}
	\hfil
	\subfloat[Plants occlusion]{\includegraphics[width=1in,height=1.4in]{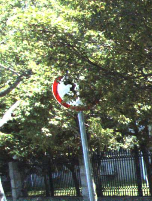}%
	\label{fig_PicPlanOcclusion}}
	\hfil
	\subfloat[Tilted signs]{\includegraphics[width=1in,height=1.4in]{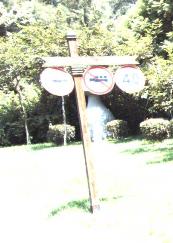}%
	\label{fig_PicTilted}}
	\hfil
	\subfloat[Wrong angle]{\includegraphics[width=1in,height=1.4in]{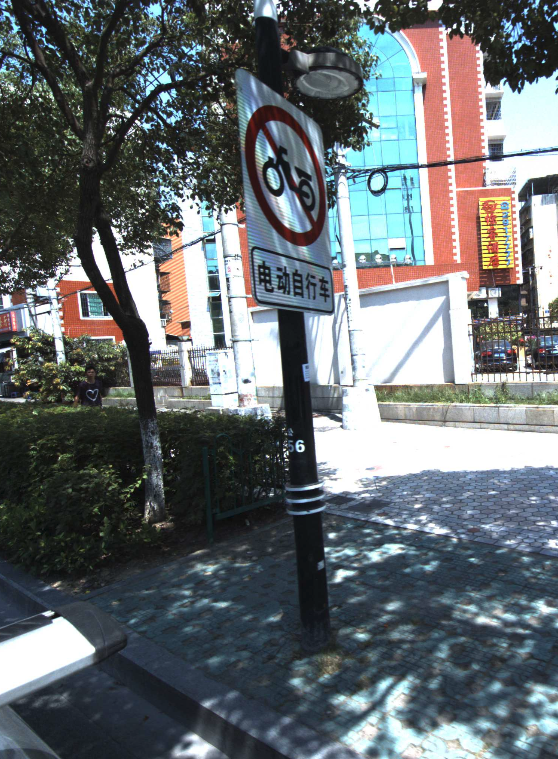}%
	\label{fig_PicWrongAngle}}
	\hfil
	\subfloat[Wrong height]{\includegraphics[width=1in,height=1.4in]{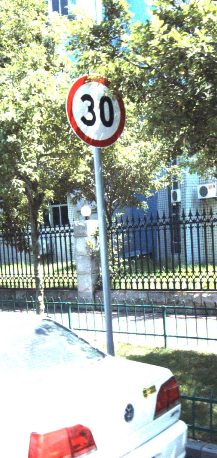}%
	\label{fig_PicWrongHeight}}
	\caption{Traffic signs with low visual recognizability.}
\label{fig_sim}
\end{figure}

Although promising results have been achieved in the areas of traffic sign detection and classification \cite{zhu2016traffic, soilan2016traffic, yu2016bag}, so far, only a few works have focused on computing the visibility of traffic signs from a given viewpoint for smart driver assistance systems or transportation facilities maintenance purposes. Of those works, most of them are based mainly on sign images and videos using computer vision methods \cite{gonzalez2011automatic, doman2014estimation, khalilikhah2016analysis}; some works research the occlusion of traffic signs based on point clouds acquired by mobile laser scanning (MLS). With the limitation of lighting conditions, and a fixed viewpoint position and view angle of getting images from cameras, there is no way to get all the information to compute a traffic sign's visual recognizability from any position above road surface. Therefore, it is not feasible to use the image or video to calculate the visual recognizability in any position of the traffic sign. The development of MLS technology makes it possible to evaluate the traffic sign's visual recognizability from 3D measurable point clouds. Unlike optical imaging, in addition to taking pictures, MLS can provide a complete point cloud of the entire roadway scene without limitations of lighting conditions. One can extract all of the information from a point cloud needed to compute a traffic sign's visual recognizability. Its appearance provides a new way to research visibility and recognizability in a measurable and real traffic environment. Fig. 1 illustrates some examples of traffic signs that with low visual recognizability caused by object occlusion, plants occlusion, tilt, wrong depress angle and height, respectively. Fig. \ref{fig_PicObjectOcclustion} and \ref{fig_PicObjectOcclustion2} are the same place from different viewpoints above the road. Of course, the height influence for traffic sign visibility is not obvious, but the influences it brings do exist.

In this paper, we present an automatic traffic sign timely visual recognizability evaluation model based on human visual cognition theory using 3D measurable point clouds acquired by an MLS system. We summarize our main contributions as follows:

\begin{enumerate}[]
	\item We addressed the conception of a visibility field to reflect the visible distribution in a 3D space. For a viewpoint in a 3D space, we presented a VEM model based on human visual cognition theory to compute a traffic sign's visibility. The VEM combines the principle of retinal imaging with the actual human driving situation to measure the clarity of traffic signs for a given viewpoint.
	\item Comparing with the VEM, we addressed the conception of a visual recognizability field to reflect visual recognizable distribution in a 3D space. For a viewpoint in a 3D space, we established a VREM model to measure a traffic sign's visual recognizability for a given viewpoint. 
	\item To evaluate a traffic sign's visual recognizability, we propose the TSTVREM model by combining the VREM, the actual maximum continuous visual recognizable distance, and traffic big data.
	\item We presented an automatic algorithm to realize the TSTVREM. It includes extracting a traffic sign point clouds, segmented surrounding point clouds in front of traffic signs on the right of roadway, and computing viewpoints according the different lanes based on extracted road marking point clouds.
\end{enumerate}

\begin{figure*}[!t]
	\centering
	\includegraphics[width=5.5in,height=3.5in]{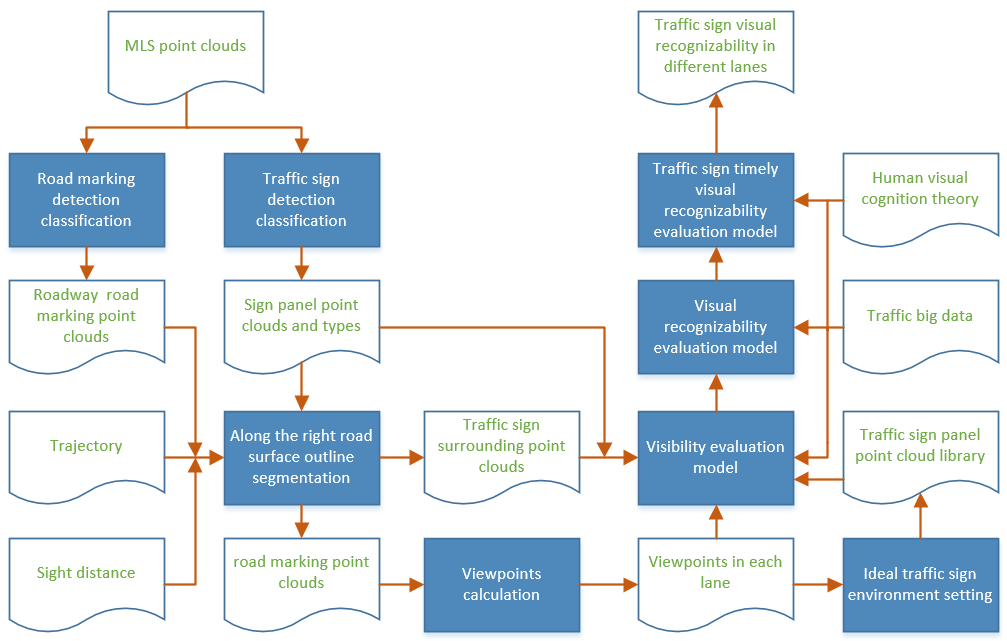}
	\caption{Pipeline of traffic sign timely visual recognizability evaluation.}
	\label{img_framework}
\end{figure*}

The pipeline of traffic sign timely visual recognizability evaluation is illustrated in Fig. \ref{img_framework}. Firstly, we segment MLS point clouds into ground point clouds and nonground point clouds and extract the traffic sign point clouds from nonground point clouds. Secondly, we split surrounding point clouds in front of the traffic sign from nonground point clouds according to the designed visual cognition distance of the roadway. Thirdly, we extract the road marking point clouds from the ground point clouds and then get the viewpoints in different lanes. Finally, we use our algorithm to realize the TSTVREM model based on the extracted point clouds and viewpoints to get the visual usability of a traffic sign.

This paper is organized as follows. A review of the previous work is given in Section \uppercase\expandafter{\romannumeral2}. Sections \uppercase\expandafter{\romannumeral3} and \uppercase\expandafter{\romannumeral4} describe the definition of the TSTVREM model and its implementation, respectively. Section \uppercase\expandafter{\romannumeral5} shows experiments and Section \uppercase\expandafter{\romannumeral6} concludes this paper.

\section{Related Work}
With the rapid development of laser radar, especially the application of MLS systems that are able to collect accurate and reliable 3D point clouds, these point clouds provide geometric and radiometric information for infrastructure facilities. This makes it more simple and efficient for people to survey an urban or a roadway environment. Wen et al. \cite{wen2016spatial} state the attributes of the MLS system and its data acquisition process. According to the main focus of this paper, we divide the related work into three categories: traffic sign detection and classification, road marking detection and classification, and visibility research. 

\subsection{Traffic Sign Detection and Classification}
The goal of traffic sign detection and classification is to find the locations and types of traffic signs. Most of the existing traffic sign detection and classification methods are based on extracting color and shape information from images or videos. The color-based method mainly uses the color space design to segment the sign candidate area, and then uses the shape feature or the edge feature to extract the traffic sign \cite{marinas2011detection,lillo2015traffic}. The shape-based detection methods include: shape matching \cite{li2015novel}, Hough transform\cite{qin2010unified}, HOG feature and SVM classification \cite{greenhalgh2012real}, HOG feature and Convolutional Neural Networks (CNNs) \cite{yang2016towards}, among others. However, the detection performance of these methods is heavily effected by weather conditions, illumination, view distance and occlusion.

Recently, researchers have developed various methods to detect traffic signs in point clouds. Yang et al.\cite{yang2015hierarchical} proposed a method to extract urban objects (include traffic signs) from MLS over-segmentations of urban scenes based on supervoxels and semantic knowledge. Lehtom et al. \cite{lehtomaki2016object} took spin images and LDHs into account to recognize the objects (including traffic signs) in a roadway environment using a machine learning method. Wen et al. \cite{wen2016spatial} presented a spatial-related traffic sign detection process for inventory purpose. Some researchers also proposed the detected method combining a 3D point cloud and 2D  images for getting a good detecting performance \cite{yu2016bag,soilan2016traffic,tan2016weakly,ai2016automated}. 

\subsection{Road Marking Detection and Classification}
Road markings on paved roadways, as critical features in traffic management systems, have important functions in providing guidance and information to drivers and pedestrians. Guan et al. \cite{guan2014using} proposed an algorithm to extract road markings using the point-density-dependent multi-threshold segmentation and morphological closing operation. \cite{guan2015using},\cite{soilan2017segmentation} mentioned a method for rapidly extracting road marking by generating 2D georeferenced images from 3D point clouds. The method of extracting road marking by converting 3D point clouds into 2D georeferenced feature images will lead to incompleteness and incorrectness in the feature extraction process, Yu et al. \cite{yu2015learning} extracted the road marking directly from the 3D point clouds relying on the reflective properties and classified them using deep learning. 

\subsection{Visibility Research}
Doman et al. \cite{doman2010estimation} proposed a visibility estimation method for traffic signs as part of nuisance-free driving safety support systems by preventing the provision of too much information to a driver. They improved their method by considering temporal environmental changes in \cite{doman2011estimation} and integrated both the local and global features in a driving environment with \cite{doman2011estimation} in \cite{doman2014estimation}. They use different contrast ratio and distance counted by pixel numbers in different area of a image to compute visibility of traffic sign. This way is limited by the position of viewpoint and weather condition and they did not consider the traffic sign placement, occlusion, road curve, and subjective factor mentioned in Section \uppercase\expandafter{\romannumeral1} in their model. 

Katz et al. proposed a Hidden Point Removal (HPR) operator to get visible points from a given viewpoint \cite{katz2007direct}, applied it for improving the visual comprehension of point sets in \cite{katz2013improving}, and researched what properties should transformation the function of an HPR operator satisfy in \cite{katz2015visibility}. Based on the HPR operator, traffic signs occlusion detection from a point cloud has been researched in \cite{huang2017traffic}. This paper measured the occlusion by occluded distribution index and occlusion gradient index. However, other factors including occluded area proportion, the influence of driver speed on vision, road curve, number of lanes, and others have not been considered. Those factors are important for a traffic sign and influence the visibility and recognizability of the traffic sign. Beside this, the HPR operator only can detect the surface points which occluded the traffic sign, and it cannot detect all the occlusion point clouds when the occlusion point cloud is composed by many objects or plans. 

\section{Definition of TSTVREM Model}
\begin{figure}[!t]
	\centering
	\includegraphics[width=3.5in]{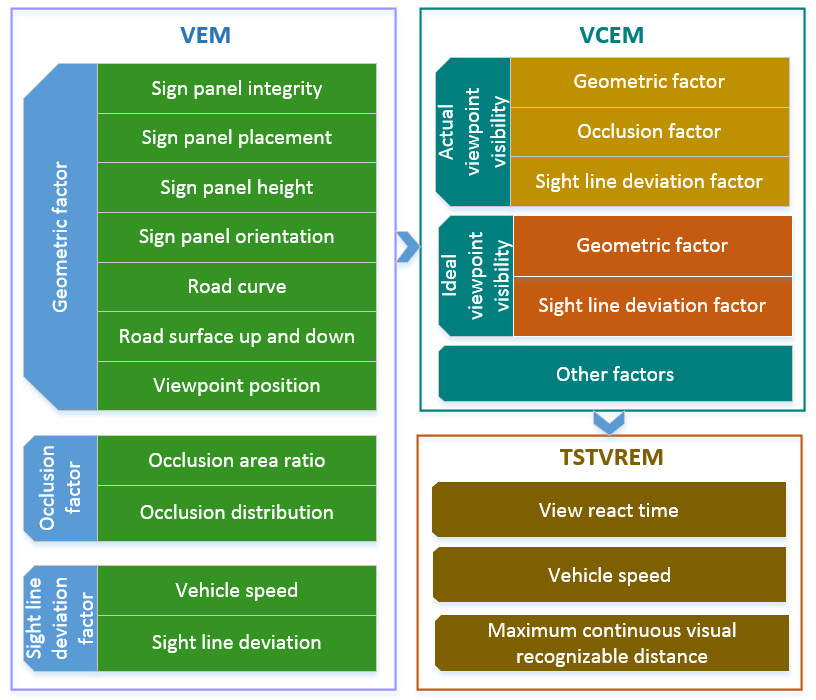}
	\caption{The framework of TSTVREM model.}
	\label{img_model}
\end{figure}

The framework of TSTVREM model as shown in Fig. \ref{img_model}. In this paper, the phrase “\textit{viewpoint visibility}” means the visibility of a traffic sign for a given viewpoint, and the phrase “\textit{viewpoint recognizability}” means the recognizable degree of a traffic sign for a given viewpoint. The framework can be divided into three parts: visibility field definition and VEM model (Section \uppercase\expandafter{3.1}), visual recognizability field definition and VREM model (Section \uppercase\expandafter{3.2}), and traffic sign timely visual recognizability evaluation (Section \uppercase\expandafter{3.3}).

\subsection{Definition of Visibility Field and VEM Model}

\begin{figure*}[!t]
	\centering
	\subfloat[viewpoints]{\includegraphics[width=2in,height=2in]{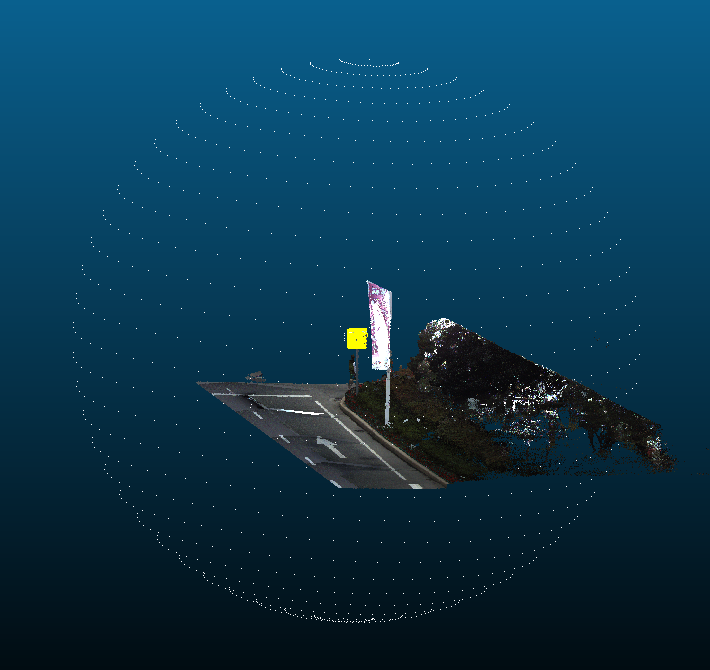}%
		\label{figvisualField}}
	\hfil
	\subfloat[Actual visibility field]{\includegraphics[width=2in,height=2in]{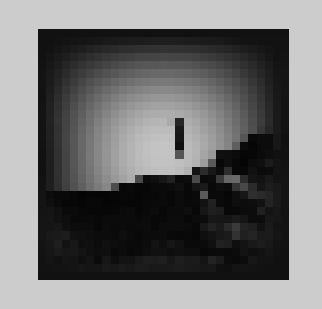}%
		\label{visibilityFieldOccluded}}
	\hfil
	\subfloat[Ideal visibility field]{\includegraphics[width=2in,height=2in]{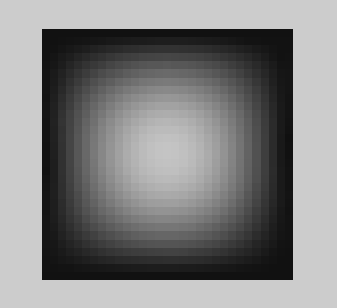}%
		\label{figvisibilityFieldNoOccluded}}
	\caption{Hemispherical visibility field of a traffic sign.}
	\label{figVisibilityField}
\end{figure*}

Visibility field definition: for a given 3D environment around a target object, the visibility of each viewpoint in 3D space around the object constitutes a visibility field. It reflects the visible distribution about a target object in 3D space. The visibility field can be divided into the actual visibility field and ideal visibility field according to an actual traffic sign set in a real road environment and its corresponding ideal traffic sign in an ideal road environment, respectively. Take a traffic sign as an example; the hemispherical visibility field for the traffic sign is shown in Fig.\ref{figVisibilityField}; the environment with occlusion or not is the actual visibility field or ideal visibility field separately; the traffic sign is yellow in \ref{figvisualField}; white indicates visibility 1 and black indicates visibility 0 in Fig. \ref{visibilityFieldOccluded} and \ref{figvisibilityFieldNoOccluded}.

\subsubsection{Actual traffic sign visibility field and VEM model}

For a traffic sign in an actual traffic environment, the visibility of each viewpoint, which has a fixed height above on the road surface of driving direction lanes in front of the traffic sign, constitutes an actual traffic sign visibility field. It is related to the orientation of traffic sign's panel, traffic sign's height, observation distance, road curve, road surface up and down, viewpoint position, degree of occlusion, and sight line deviation, among others. We call the influence of traffic sign's panel orientation, traffic sign's height, observation distance, road curve, road surface up and down, viewpoint position as geometric factor. The VEM model is constructed by geometric factor $E^{geo}$, occlusion factor $E^{occ}$, and sight line deviation factor $E^{sight}$.
The viewpoint visibility of a traffic sign $E^{visibility}$ can be defined as follows:
\begin{equation} \label{visibilitySimple}
E^{visibility}=E^{geo}*E^{occ}*E^{sight}
\end{equation}

In the following part, we describe how to evaluate the geometric factor, occlusion factor, and sight line deviation factor, respectively.
\begin{itemize}
	\item{Geometric factor evaluation
		
		In order to make the visibility calculation consistent with the human visual recognition theory. We use the principle of retinal imaging to consider the impact of geometric factor. The evaluation of geometric factor $E^{geo}$ is given as blow. 
		
		\begin{equation} \label{geometric}
		E^{geo}=A^{view}/A_{type}^{standard}
		\end{equation}
		
		\begin{itemize}
			\item[-] $A^{view}$ : the retinal imaging area of the traffic sign observed from a given viewpoint.
			\item[-] $A_{type}^{standard}$: the standard retinal imaging area of its corresponding ideal traffic sign for a given viewpoint, which is at the normal of ideal traffic sign panel and passes through the center of the panel and has a fixed standard distance $d^{standard}$ to the panel. In order to make $E^{geo} \leq 1$, we take the standard distance $d^{standard}$ less than $3$ m. Because the traffic sign has disappeared in driver's view field, there is no meaning to compute visibility of a traffic sign when the observation distance less than $d^{standard}$ meter. Different types of traffic signs have different standard retinal imaging areas.
	\end{itemize}}
	
	Obviously, $E^{geo}$ is inversely proportional to the angle between the line connecting the viewpoint to the traffic sign panel center and the normal passing through the traffic sign panel center (orientation factor), observation distance, and height difference between the viewpoint and traffic sign panel center.
	
\end{itemize}

\begin{itemize}
	\item{Occlusion factor evaluation
		
		We also use the principle of retinal imaging to consider the impact of the occlusion area ratio. Apart from that, we introduce the occlusion distribution factor as \cite{huang2017traffic} did. The evaluation of the occlusion degree $E^{od}$ is given as blow.   		
		\begin{equation} \label{occlusiondegree}
		E^{od}=\alpha * \frac{A^{occ}}{A^{view}}+ \beta *(1-\frac{\| c^{occ}-c^{sign} \|}{d^{max}})*\frac{A^{occ}}{A^{view}}
		\end{equation}
		\begin{itemize}
			\item[-] $\frac{A^{occ}}{A^{view}}$: the retinal imaging occluded area ratio. $A^{occ}$ is the retinal imaging area of the occluded traffic sign region for a given viewpoint.    
			\item[-] $1-{\| c^{occ}-c^{sign} \|}/{d^{max}}$: the occlusion distribution. $\| c^{occ}-c^{sign} \|$ is the distance between the center point of occluded traffic sign region $c^{occ}$ and traffic sign panel center point $c^{sign}$. $d^{max}$ is the maximum length from $c^{sign}$ to each vertex of the boundary polygon of the traffic sign panel.
			\item[-] $\alpha$, $\beta$: weights. They should satisfy the condition: $\alpha+\beta=1$.
	\end{itemize}}
	After adding the punishment item $\lambda$, the evaluation of the occlusion factor $E^{occ}$ is given as below. 
	\begin{equation} \label{occlusion}
	E^{occ}=e^{-\lambda * E^{od}}
	\end{equation}
	In order to make sure $E^{occ}$ nearly to zero when $E^{od}$ is nearly one, $\lambda$ should meet the conditions $\lambda \geq 6$. Obviously, when the degree of occlusion $E^{od}$ is constant, $E^{occ}$ decreases as the penalty parameter $\lambda$ increases; when the penalty parameter $\lambda$ is constant, $E^{occ}$ decreases as the degree of occlusion $E^{od}$ increases. Therefore, Formula \ref{occlusion} meets our expectations that the effect on visibility decreases when the degree of occlusion increases.
\end{itemize}

\begin{itemize}
	\item{Sight line deviation factor evaluation
		
		For an object we see in a fixed viewpoint, it is more clear we look at it in the front view than we when look at it in an oblique view. This factor reflects how the different imaging positions on the retina may lead to different visibilities. Furthermore, a driver on the road at different drive speeds will lead to him having different GFOVs \cite{parkes2010geometric}. Combine with those above, the sight line deviation factor evaluation $E^{sight}$ is established as below. 	
		\begin{equation} \label{sightLine}
		E^{sight}=\left\{  
		\begin{array}{lr}  
		1, & V^{a}<0.5*V^{f} \\  
		e^{-\eta * \frac{V^{a}-0.5*V^{f}}{0.5*V^{f}}}, & 0.5*V^{f} \leq V^{a} \leq \pi/2 \\
		0, & V^{a} > \pi/2  
		\end{array}  
		\right.  
		\end{equation}
		
		\begin{itemize}
			\item[-] $V^{a}$ : sight line deviation angle between the line of sight and the line connecting the viewpoint to $c^{sign}$.
			\item[-] $V^{f}$: the actual GFOV. It depends on the actual $85^{th}$-percentile speed $v^{85}$ \cite{adminstration2009manual} which comes from traffic big data. $V^{f}$ can be denoted by a function with parameter $v^{85}$ as $V^{f}=f(v^{85})$.
			\item[-] $\eta$ : It is used as punishment item when sight line deviation angle is bigger than half of the actual GFOV.
	\end{itemize}}
\end{itemize}

To sum up all factors discussed above, For given a traffic sign and a $j^{th}$ viewpoint $p_j$ in the $i^{th}$ lane $l_i$, its visibility $E_{i,j}^{visibility}$ equals:

\begin{equation}\label{visibility}
\begin{split}
&E_{i,j}^{visibility} = E_{i,j}^{geo} * E_{i,j}^{occ} * E_{i,j}^{sight} \\ 
&= \frac{A_{i,j}^{view}}{A_{type}^{standard}}  *
e^{-\lambda *( {\alpha * \frac{A_{i,j}^{occ}}{A_{i,j}^{view}}+ \beta *(1-\frac{\| c^{occ}-c^{sign} \|}{d^{max}})*\frac{A^{occ}}{A^{view}}})}
		* E_{i,j}^{sight}
\end{split}
\end{equation}	
\subsubsection{Traffic Sign Ideal Visibility Field and VEM model}

For a traffic sign in an ideal traffic environment, the visibility of each viewpoint that has a fixed height above the surface of driving direction lanes in front of the traffic sign constitutes an ideal traffic sign visibility field.	
An ideal traffic environment is an environment that has an ideal traffic sign installed specified in a suitable placement beside the straight horizontal roadway according to traffic design installation rules, and meets the condition that there is no other object around the roadway in addition to the traffic sign. Therefore, the VEM model in an ideal environment degenerates into the product of the geometric factors $E^{geoI}$ and sight line deviation factors $E^{sightI}$. The formula for viewpoint visibility in an ideal traffic environment $E{^{visibilityI} _{i,j}}$ is shown below:

\begin{equation}\label{idealVisibility}
E_{i,j}^{visibilityI} = E{^{geoI} _{i,j}}* E_{i,j}^{sightI}  
\end{equation}	 	
\begin{itemize}
	\item[-] $E{^{geoIdeal} _{i,j}}$: the evaluation of geometric factor in an ideal traffic environment.    
	\item[-] $E_{i,j}^{sightIdeal}$: the evaluation of sight line deviation factor in an ideal traffic environment.	
\end{itemize}

\begin{equation}\label{idealVisibility}
E{^{geoI} _{i,j}} = A_{i,j}^{viewI}/A_{type}^{standard}
\end{equation}	 

\begin{itemize}
	\item[-] $A_{i,j}^{viewI}$: the retinal imaging area of the traffic sign observed from a given viewpoint in an ideal traffic environment.  
	
\end{itemize}

\begin{equation} \label{idealSightLine}
E^{sightI}=\left\{  
\begin{array}{lr}  
1, & V^{aI}<0.5*V^{fI} \\  
e^{-\eta * \frac{V^{aI}-0.5*V^{fI}}{0.5*V^{fI}}}, & 0.5*V^{fI} \leq V^{aI} \leq \pi/2  \\
0, & V^{aI} > \pi/2 
\end{array}  
\right.  
\end{equation}
\begin{itemize}
	\item[-] $V^{aI}$ : sight line deviation angle between the line of sight and the line connected viewpoint to $c^{sign}$ in an ideal traffic environment.
	\item[-] $V^{fI}$ : the ideal GFOV. It depend by the design speed $v^{design}$ of the road and $V^{fI}=f(v^{design})$.
	\item[-] $\eta$ : It is used as punishment item when sight line deviation angle big than half of the ideal GFOV.
\end{itemize}
\subsection{Definition of Visual Recognizability Field and VREM Model}
Although the VEM model uses retinal imaging area to estimate viewpoint visibility in accordance with the natural recognition process, it is still a difficult problem to determine the viewpoint recognizability only by viewpoint visibility. For example, for two viewpoints have the same visibility to the same traffic sign, the near viewpoint with occlusion, the far viewpoint without occlusion. The sign cannot be recognized from the near viewpoint, for too much effective information has been lost to occlusion, while the sign may be recognized from the far viewpoint from the blurred silhouette. In order to conquer this problem, we introduced corresponding viewpoint visibility in the ideal traffic environment as the standard to evaluate viewpoint recognizability. Beside this, the advantage of introducing ideal viewpoint visibility is that it can be used as the standard to judge whether traffic signs meet the design and installation specifications or not, and then evaluate whether the traffic sign is recognizable or not from a given viewpoint. 

Visual recognizability field definition: for a given 3D environment around a target object, the visual recognizability of each viewpoint in 3D space around the object constitutes a visual recognizability field. It reflects the visual recognizable distribution about an object in 3D space. 

In the VREM model, the viewpoint recognizability is related to the actual viewpoint visibility and the corresponding ideal viewpoint visibility, and the other factors $E^{other}$ mentioned in Section \uppercase\expandafter{\romannumeral1}. We denote the intersection point that the polyline made up by viewpoints intersects with the perpendicular line to the driving direction and passing through the traffic sign center as $p^{intersect}$. Remember the intersection point which the line perpendicular to driving direction and passing traffic sign center intersects with the right road marking outline with $p^{rOutline}$. Remember the distance from viewpoint to $p^{intersect}$ along the polyline made up by viewpoints and the distance from $p^{intersect}$ to $p^{rOutline}$ with $d^{length}$ and $d^{width}$ respectively. The corresponding viewpoint in an ideal traffic environment has the same $d^{length}$ and $d^{width}$ as the viewpoint in an actual traffic environment. 

The visual recognizability $E_{i,j}^{recognizability}$ is given as follows:

\begin{equation}\label{cognition}
E_{i,j}^{recognizability}=\textbf{1}\{\gamma * \frac{E_{i,j}^{visibility}}{E_{i,j}^{visibilityI}} + \delta *E^{other} > \sigma \}
\end{equation}
\begin{itemize}
	\item[-] $\gamma,\delta$ : weights. $\gamma$ and $\delta$ meet the condition $\gamma + \delta =1$. In this paper, we leave the research of other factors influence to cognition for the future, that is $\delta =0$. 
	\item[-] $\sigma$ : threshold. It is used to judge whether the traffic sign at a viewpoint is can be recognized or not. 
\end{itemize}
\subsection{Traffic Sign Timely Visual Recognizability Evaluation}
According to the Manual on Uniform Traffic Control Devices (MUTCD in the United States) \cite{adminstration2009manual} or Road Traffic Signs and Markings (GB 5768-1999 in China) \cite{yang1999gb}, Sight Distance (SD) is a length of road surface from the point which a driver can see traffic sign with an acceptable level of clarity to the traffic sign. SD $d^{sightDistance}$ is given based on a driver's ability for visual recognition under a designed vehicle speed. We use the visual recognizability field composed by the viewpoints within SD length of forward direction area of road surface to evaluate visual recognizability of a traffic sign. 

The traffic sign timely visual recognizability is evaluated according to different lanes in the forward direction area of a road surface. It is not only related to the actual maximum continuous visual recognizable distance $d_{i}^{maxCogLength}$, which the maximum continuous length of the viewpoints can be recognized along a lane, but also to the actual vehicle speed on the roadway and VRT. VRT is simply the time necessary for a driver to detect, read, and react to the message displayed on an approaching on-premise sign that lies within his or her cone of vision \cite{bertucci2006sign}. Once VRT is ascertained, VRD $d^{vrd}$ for a given sign location, or the distance which a vehicle travels during the VRT interval, can be calculated; therefore, $d^{vrd}=v^{85}*t^{vrt}$. The relationship between $d_{i}^{maxCogLength}$ and $d^{vrd}$ determines whether the traffic sign has enough time to recognize or not. The evaluation of traffic sign visual recognizability $E_i^{recognizability}$ is given as below. \textbf{1}\{condition\} return 1 when condition is ture, else return 0.   		
\begin{equation}\label{maxLength}
\begin{split}
E_i^{recognizability} &= \textbf{1}{\{ d_{i}^{maxCogLength} \geq d^{vrd} \}}\\
				&=\textbf{1}{\{ d_{i}^{maxCogLength} \geq v^{85}*t^{vrt} \}}
\end{split}
\end{equation}


\section{TSTVREM Model Implementation}
\begin{figure*}[!t]
	\centering
	\subfloat[Actual environment]{\includegraphics[width=3.5in]{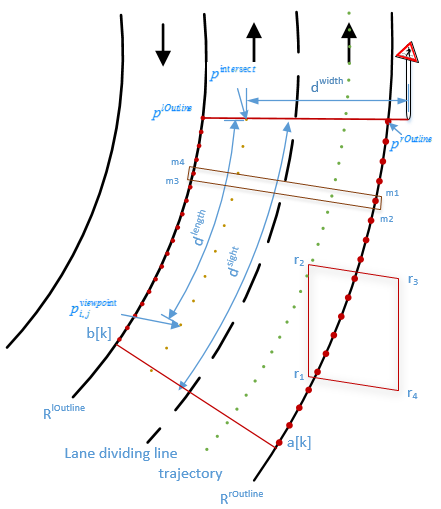}%
		\label{img_actualEnvironment}}
	\hfil
	\subfloat[Ideal environment]{\includegraphics[width=3.4in]{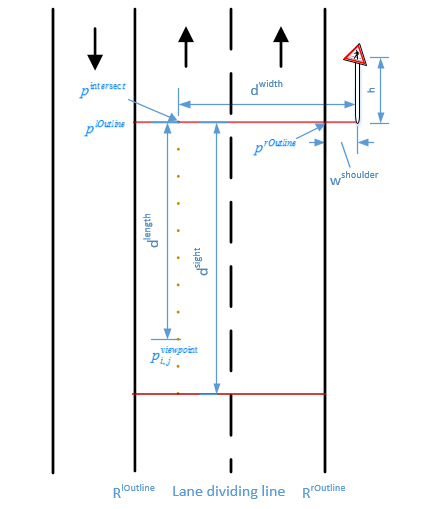}%
		\label{img_idealEnvironment}}
	\caption{Illustration of TSTVREM model implementation.}
	\label{img_realizeTSTVREM}
\end{figure*}

We construct an automatic algorithm to implement the TSTVREM model in point clouds acquired by an MLS system. The input is a road's point clouds and its trajectory, the output is a visibility field, recognizability field, and traffic sign timely visual recognizability in each lane. First, we detect and classify the traffic signs and extracted road marking from the input point clouds. This step is called preliminary work. Second, we abbreviate the surrounding point clouds on the right side above the roadway in front of a traffic sign as traffic sign surrounding point clouds and segment traffic sign surrounding point clouds and road marking point clouds along the roadway according to SD for each traffic sign. This step is called the segment process. Then, the viewpoints position above road surface forward direction region is computed form the segmented road marking point clouds. This step is called viewpoints computing. In the end, we use the traffic sign panel point cloud, the traffic sign surrounding point clouds, and the viewpoints together to extract information to compute traffic sign timely visual recognizability using TSTVREM model. This step is called traffic sign timely recognizability computing. The following part is to introduce these processes separately. All symbols used to describe the TSTVREM model implementation are illustrated in Fig. \ref{img_realizeTSTVREM}. 

\subsection{Preliminary Works}
For input point clouds of a road, we adopt the method proposed by Wen et al. \cite{wen2016spatial} for traffic sign detection and classification. The output is all a traffic sign's panel point clouds and its type. Using the sign type and speed limit of the roadway, we can get the SD of each traffic sign $d_{type}^{sightDistance}$ according to the country's traffic signs design specifications. The extracted traffic signs' panel point clouds combine with SD are used to segment traffic sign surrounding point clouds.

We adopt the algorithm presented by Yu et al. \cite{yu2015learning} for road markings detection and classification. The extracted road markings are segmented by extracted traffic signs position and SD for every traffic signs. The type of road markings is used to distinguish road surface forward direction region and different lanes for each segmented road marking point clouds.

\subsection{Segment Process}
Using the traffic sign panel point clouds, we can compute the traffic sign center $c^{sign}$ and find the nearest trajectory point $p^{traj}$ to $c^{sign}$. The vector from $c^{sign}$ to $p^{traj}$ is denoted as $\textbf{h}^{sign}$. 

From the road markings point clouds we extracted, we use the different length of clusters along the roadway direction to distinguish solid and dashed lines. Its length threshold is the smallest SD $30$ meters in \cite{yang1999gb}. If the solid line is not continuous or partially missing because of the low reflectivity, we use its attribute that it approximately parallels with trajectory line to complete it. For solid lines in the right of $p^{traj}$, we sorted the distances from solid line to $p^{traj}$ perpendicular to the driving direction. The solid line which has the maximum distance is the right roadside outline $\textbf{R}^{rOutline}$. For solid lines in the left of $p^{traj}$, we sorted distance from the solid line to $p^{traj}$ perpendicular to the driving direction. The solid line which has the nearest distance is the left outline $\textbf{R}^{lOutline}$ of the road surface forward direction region. The vector $\textbf{h}^{sign}$ intersects $\textbf{R}^{rOutline}$ with point $p^{rOutline}$. The vector $\textbf{h}^{sign}$ intersects $\textbf{R}^{lOutline}$ with point $p^{lOutline}$. Remember the distance from $p^{lOutline}$ to $p^{rOutline}$ with $d^{drivingWidth}$. 

If the road surface forward direction region is not detected because of road markings wear and tear or no road marking, we get $\textbf{R}^{lOutline}$ and $\textbf{R}^{rOutline}$ through left and right move the trajectory and minus the height of the MLS device for every trajectory point.

The method to get $p^{lOutline}$ and $p^{rOutline}$ from solid lines is as follows. In the $xOy$ plane, select the nearest point $t_k$ to the traffic sign in the trajectory. The two points near the $t_k$ in trajectory are denoted by $t_{k-1}$ and $t_{k+1}$. Segment the solid lines in a slice along the $\textbf{h}^{sign}$ and compute the centers of the sliced point clouds for every cluster. The $p^{lOutline}$ and $p^{rOutline}$ is selected by the distance and side from centers to line $t_{k-1}$ to $t_{k+1}$. 

Using $p^{lOutline}$ and $p^{rOutline}$ as start points, constantly cutting pieces of road marking clusters along the trajectory by interval \cite{guan2014using} and remember the intersection with $b[]$ and $a[]$ respectively, compute the center of $b[]$ and $a[]$ denoted as $m[]$, until the distance accumulated in $m[]$ is bigger than the traffic sign's SD. Remember the point which is in the line from last point to second last point in $m[]$, and its position at last point minus the extra length longer than SD as $p^{tmp}$. The last point in $b[]$ and $a[]$ should be adjustment by the ratio of $p^{tmp}$ in line from last point to second last point in $m[]$.

We construct two rectangles in a vertical plane to segment out the traffic sign surrounding point clouds and solid road marking point clouds. The rectangles have a horizontal edge always perpendicular to $a[]$, and their horizontal length and vertical height of edges can change according to our will. The octree segment method is used to segment out the traffic sign environment point cloud and solid road marking point cloud along $a[]$. The vector from $a[k]$ to $a[k+1]$ denoted as $\textbf{a}^{kk+1}$, the horizontal unit vector perpendicular to $a[]$ can be denoted as $\textbf{h}^{kk+1}=(\textbf{a}^{kk+1} \times \textbf{z})/\Arrowvert \textbf{a}^{kk+1} \times \textbf{z}  \Arrowvert$. Remember the coordinate of $a[k]$ is $a[k]=(x_k,y_k,z_k)$. The four coordinates of the rectangle used to segment out the traffic sign environment point cloud we constructed are $r_1,r_2,r_3,r_4$, then $r_1=(x_k,y_k,z_k+0.3)$, $r_2=(x_k,y_k,z_k+3)$, $r_3=r_2+2*\textbf{h}^{kk+1}$, and $r_4=r_1+2*\textbf{h}^{kk+1}$. We use the same method to get the rectangle for segment solid road markings. The four coordinates of the rectangle used to segment out the road marking point cloud we constructed are $m_1,m_2,m_3,m_4$, then $m_1=(x_k,y_k,z_k+1)$, $m_2=(x_k,y_k,z_k-1)$, $m_3=m_2-[{\Arrowvert p^{lOutline} - p^{rOutline} \Arrowvert}+0.1]*\textbf{h}^{kk+1}$, $m_4=m_1-[{\Arrowvert p^{lOutline} - p^{rOutline} \Arrowvert }+0.1]*\textbf{h}^{kk+1}$.

\subsection{Viewpoints Computing}

In order to reduce the effect of reflectivity on the extraction of traffic marks, we use solid road marking lines to calculate the number of lanes. If the lane standard width is denoted as $d^{lane}$, then we get the number of lanes $m=INT(d^{drivingWidth}/d^{lane})$. The actual lane width $d^{laneWidth}$ is computed by formula $d^{laneWidth}=d^{drivingWidth}/m$.

The next step is to get all the lane dividing lines between $\textbf{R}^{rOutline}$ and $\textbf{R}^{lOutline}$. We denote the dividing lines as $\textbf{R}_1^{Lane}, \textbf{R}_2^{Lane}, ..., \textbf{R}_i^{Lane}, ..., \textbf{R}_{m-1}^{Lane}$ from right to left, $\textbf{R}^{rOutline}$ as $\textbf{R}_0^{Lane}$, $\textbf{R}^{lOutline}$ as $\textbf{R}_m^{Lane}$. The $\textbf{R}_0^{Lane}$ and the $\textbf{R}_m^{Lane}$ is known and depressed by arrays $a[]$ and $b[]$ respectively. The unit vector $a[k]$ to $b[k]$ is be denoted by $\textbf{h}_k^{ab}=(b[k]-a[k])/{\| b[k]-a[k] \|}$ firstly. Then we get the split points in lane dividing lines. The $k^{th}$ point $p_{i,k}$ in $\textbf{R}_i^{Lane}$ is $p_{i,k}=p_{i-1,k} + d^{laneWidth} * \textbf{h}_k^{ab}$ and $p_{0,k}=a[k]$.

\begin{figure}[!t]
\centering
\subfloat[Top view]{\includegraphics[width=1.3in,height=1.6in]{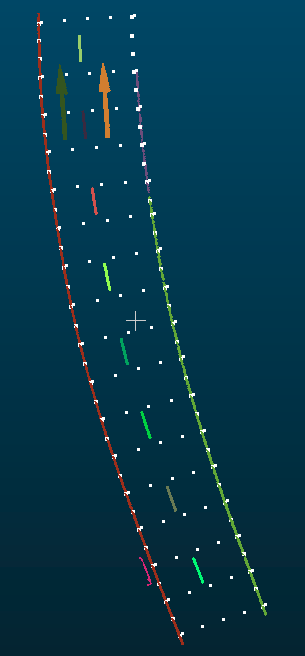}%

\label{fig_first_case}}
\hfil
\subfloat[Side view]{\includegraphics[width=1.3in,height=1.6in]{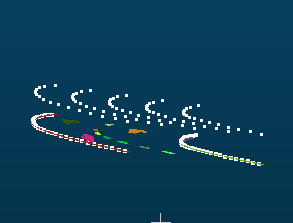}%
\label{fig_second_case}}
\caption{Viewpoints computation result.}
\label{figViewPoints}
\end{figure}

When we get every lane dividing line expressed in arrays, we can use interpolation or a sampling method as above to get any point in a lane. As the method interpolation or sampling is determined, we got a column points (maybe more column points) along a lane. The point $p_{i,j}^{col}$ is used to denote $j^{th}$ point in the column between $\textbf{R}_i^{Lane}$ and $\textbf{R}_{i+1}^{Lane}$. We get the viewpoint $p_{i,j}^{viewpointO}$ by plus observation height $h^{eye}$ to the $z$ coordinate of $p_{i,j}^{col}$. The observation height is usually set to $1.2$ meters above the road surface \cite{banks2002introduction}. The result of calculated viewpoints is shown in Fig. \ref{figViewPoints}.

\subsection{Traffic Sign Timely Visual Recognizability Computing}
Through those three steps above, for a traffic sign, we get a traffic sign panel point cloud, a traffic sign environment point cloud, and viewpoints along the lanes. The remaining work is how to use them to extract information as the input of the TSTVREM model to compute traffic sign timely visual recognizability. Its details include: compute retinal imaging area of a point cloud for a given viewpoint, get the occlusion point cloud projected in traffic sign panel, get the angle of sight line deviation, set the ideal traffic environment.

\subsubsection{Retinal Imaging Area Computing}
						
The first aspect of computing the retinal imaging area is to rotate the coordinates of point clouds to the human view. The line from the traffic sign panel point cloud center $c^{sign}$ to the viewpoint $p_{i,j}^{viewpointO}$ is denoted as $l^{cp}$. For a group of traffic sign panel point clouds, a traffic sign surrounding point cloud and a viewpoint, we rotate their z-axis coordinates to the line $l^{cp}$ using quaternions rotation method \cite{kuipers1999quaternions}, and move the origin of coordinate system to the rotated $c^{sign}$. Remember the coordinate transformed traffic sign panel center as $c^{signT}$ and $c^{signT}=(0,0,0)$. Remember the coordinate transformed $p_{i,j}^{viewpointO}$ as $p_{i,j}^{viewpoint}$, the coordinate-transformed traffic sign panel point cloud and coordinate-transformed traffic sign environment point cloud are denoted as $\textbf{S}^{sign}$ and $\textbf{S}^{environment}$, respectively. Next, project the $\textbf{S}^{sign}$ to its $xOy$ and denote it as $\textbf{S}^{signProjected}$. Then, the edges of $\textbf{S}^{signProjected}$ are computed by the alpha shape algorithm \cite{edelsbrunner1983shape}. The parameter alpha of this algorithm is remembered $d^{alpha}$. Remember the polygon composed by those edges as $e^{polygon}$. At last, we can use polygon area formula to compute edge area and map the area to retinal imaging area using human retinal imaging principle \cite{kaiser2007calculation}. The distance from the center of eye's entrance pupil to the retina $d^{retina}$ is set to $17$ millimeters.
	      
\subsubsection{Occlusion Point Cloud Obtaining}
						
Compute the distances from $c^{signT}$ to every vertex of $e^{polygon}$ and select out the vertex $v$ that has the largest distance $d^{max}$. Remember the line vector from $p_{i,j}^{viewpoint}$ to $c^{signT}$ as $\textbf{g}$ and the line vector from $p_{i,j}^{viewpoint}$ to $v$ as $\textbf{g}^{max}$. The angle $\varphi$ between $\textbf{g}$ and $\textbf{g}^{max}$ equals $\varphi = \arctan (d^{max} / \| p_{i,j}^{viewpoint}-c^{signT} \|)$. The vector from $p_{i,j}^{viewpoint}$ to a point $p_l^{environment}$ in $\textbf{S}^{environment}$ is labeled as $\textbf{g}_l$. The symbol $\tau_l$ is the angle between $\textbf{g}$ and the line $\textbf{g}_l$ and $\tau_l= \textbf{g} * \textbf{g}_l / (|\textbf{g}| * |\textbf{g}_l|)$. For every point in $\textbf{S}^{environment}$, if $\tau_l \leq \varphi$, then compute the intersection point of $\textbf{g}_l$ with xOy plane, add the intersection point into point cloud $\textbf{S}^{intersection}$ and add $p_l^{environment}$ into occlusion point cloud $\textbf{S}^{occlusion}$. For every point in $\textbf{S}^{intersection}$, if it inside the polygon $e^{polygon}$, then add it into the occluded point cloud  $\textbf{S}^{occluded}$, else delete its corresponding point in $\textbf{S}^{occlusion}$. The retinal imaging area of $\textbf{S}^{occluded}$ is computed by the alpha shape algorithm and human retinal imaging principle.

\subsubsection{Sight Line Deviation Computing}
			
The sight line of a driver in point $p_{i,j}^{viewpoint}$ is the line $\textbf{e}_{i,j}^{sight}$ defined by neighboring points from $p_{i,j}^{viewpoint}$ to $p_{i,j+1}^{viewpoint}$. The sight line deviation angle $V^{angle}$ is the angle from $\textbf{e}_{i,j}^{sight}$ to $\textbf{g}$, and $V^{angle} = (\textbf{e}_{i,j}^{sight} * \textbf{g}) / (|\textbf{e}_{i,j}^{sight}|*|\textbf{g}|)$.

\subsubsection{Ideal Traffic Environment Setting}

For all kinds of traffic signs included in a state traffic system, we need to build a traffic sign panel point cloud library of which each class of traffic signs contains one. We choose one traffic sign for each class from actual traffic environment point clouds to constitute the library. Coordinates of every traffic sign in the library are transformed such that their normal vector is parallel with y-axis and their center is the origin O.

As the traffic sign design manual of a roadway, the height $h$ above the road surface, depression angle $\phi$, the angle in the direction road users are to pass $\psi$, road shoulder width $w^{shoulder}$ are specified. The coordinate system of ideal traffic environment is set as follows. According to whether the traffic sign can be detected, it is divided into two parts. 

If the road marking is detected. We use the origin O as $p^{rOutline}$, the y-axis as $\textbf{R}^{rOoutline}$. The normal vector of the traffic sign panel $\textbf{n}^{signI}$ is $[\cos \phi * \sin \psi, \cos \phi * \cos \psi, \sin \phi]^T$. If the traffic sign in the right of roadway, the traffic sign panel center $c^{signI}$ equals $(w^{shoulder},0,h)$. If the traffic signs hanging above the roadway, coordinates of $c^{sign}$ with ($x^{sign}, y^{sign}, z^{sign})$ and $p^{rOutline}$ with $(x^{rOutline}, y^{rOutline}, z^{rOutline})$ are known, then the distance $d^{sign}$ between $(c^{sign}$ and $p^{rOutline}$ in xOy plane is computed by formula $d^{sign}=\sqrt{(x^{sign}-x^{rOutline})^2 + (y^{sign}-y^{rOutline})^2}$, coordinate of $c^{signI}$ is $(-d^{sign},0,h)$. Rotate the corresponding traffic sign panel point cloud according to $\textbf{n}^{signI}$ and $c^{signI}$ using the quaternions rotation method and coordinate translation transformation. The coordinate of the corresponding viewpoint $p_{i,j}^{viewpointI}$ of $p_{i,j}^{viewpoint}$ is $(-d_{i,j}^{width},d_{i,j}^{length},h^{eye})$. Among these, $d_{i,j}^{width}$ is the distance in xOy plane from $p_{i,j}^{viewpoint}$ to $\textbf{R}^{rOutline}$, and $d_{i,j}^{length}$ is the accumulated distance from $p_{i,0}^{viewpoint}$ to $p_{i,j}^{viewpoint}$.

If the road marking is not detected. We see the traffic sign in the ideal place. We only change the $p^{rOutline}$ coordinate into $p^{rOutline} - w^{shoulder} * \textbf{h}^{sign}$, the left work is the same as the situation of detected road marking.

\section{Experiments and Discussions}
\subsection{MLS System and Datasets}

A RIEGL VMX-450 MLS system is used in this study to acquire the datasets in the area within Xiamen Island. This system integrated two laser scanners, four high-resolution digital cameras, a GNSS, an IMU, and a DMI \cite{guan2015automated}. Two laser scanners are installed with “X’’ configuration pattern and rotate to emit laser beams with maximum valid range of 800 m at a measurement rate of 550,000 samples/s. The accuracy of scanned point cloud data is within 8 mm. Four cameras are installed in four corners to get high-resolution pictures of the surroundings. 

In order to prove the practicality of our models and algorithm both in urban roads and mountain roads, one survey including three roads were performed with the MLS to obtain the data required for this research. There are Zengcuoanbei Road (ZCABR), Longhushan Road (LHSR), and Wenping Road (WPR). The ZCABR and LHSR are urban roads, the WPR is a mountain road. The three roads information is presented in table \ref{table_twoDatasets}. The taxi travel records for the last year is extracted from the traffic big data library of Xiamen China. We use the driving speed of taxi car to estimate the actual driving speed on the road.

\begin{table}[!t]
\caption{Descriptions of the two mobile lidar datasets}
\label{table_twoDatasets}
\centering
\begin{tabular}{|c||c||c||c||c|}
\hline
\textbf{Dataset} & \textbf{Points} & \textbf{Length} & \textbf{Speed limit} & \textbf{Actual speed}\\
\hline
CABR & 131681009 & 770.06 m & 30 mph & 25.0 mph\\
\hline
LHSR & 187751087 & 1571.747 m & 40 mph & 61.1 mph\\
\hline
WPR & 130862580 & 1956.14 m & 40 mph & 42.2 mph\\
\hline

\end{tabular}
\end{table}

\subsection{Parameter Sensitivity Analysis}
For parameters in the geometric factor evaluation, once viewpoint, traffic sign, $E^{occ}$, and $E^{sight}$ are ascertained, the viewpoint visibility is inversely proportional to the size of the parameter $d^{standard}$.  

\begin{figure}[!t]
	\centering
	\includegraphics[width=3.0in]{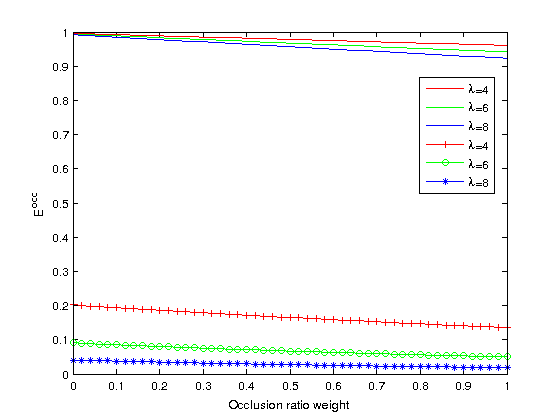}
	\caption{Occlusion value with different weights of occlusion ratio and punishment item.}
	\label{img_weightsChange}
\end{figure}

\begin{figure}[!t]
	\centering
	\includegraphics[width=3.0in]{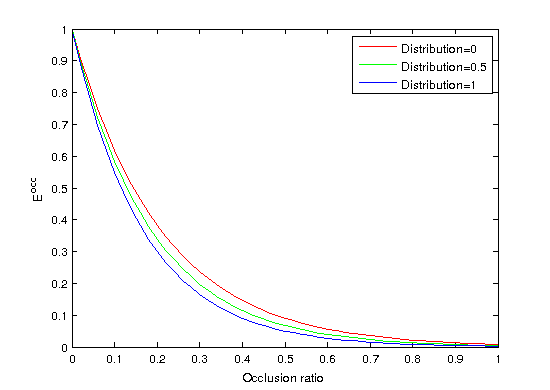}
	\caption{Occlusion value with different occlusion ratio and occlusion distribution.}
	\label{fig_occlutionRatioChange}
\end{figure}

For parameters in occlusion factor evaluation, $E^{occ} =1$ under the circumstances of no occlusion, and $E^{occ}$ nearly equals to 0 under the circumstances of half occlusion. The relationship among the parameters $\alpha$, $\beta$, and $\lambda$ in the occlusion factor evaluation part of Function \ref{visibility} is shown in Fig. \ref{img_weightsChange}. The upper three lines are generated under an occlusion ratio of 0.01 and occlusion distribution equals 0.2. From the figure, we can see that with the increment of occlusion ratio weight, $E^{occ}$ decreases gradually. This is because the occlusion distribution weight is increasing as occlusion ratio weight decreases. $E^{occ}$ decreases with the increase in $\lambda$ when other parameters ascertained and it nearly is equal to 1. Those all shown that the parameter sets do meet the demand of our model. Seemingly, to the upper three lines, the lower three lines are generated under the occlusion ratio equals 0.5 and the occlusion distribution equals 0.8, $E^{occ}$ nearly is equal to 0; this meets our model demand too. Fig. \ref{fig_occlutionRatioChange} shows that $E^{occ}$ decreases when the occlusion ratio increases and occlusion distribution near to the center of traffic sign under conditions $\alpha=0.8$, $\beta=0.2$, and $\lambda=6$.

\begin{figure}[!t]
	\centering
	\includegraphics[width=3.0in]{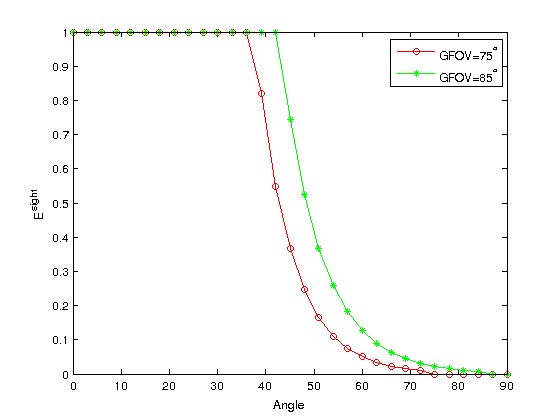}
	\caption{Sight line deviation value with different angle}
	\label{fig_sightLineDeviation}
\end{figure}

These parameters were used to test in the three datasets as Table \ref{table_parametersSet}. The $d^{length}$ in this table is the viewpoint interval along the road, its unit is the number of trajectory points. It’s about 2 m when trajectory points number equals 40 at vehicle speed 40 mph. Some research \cite{parkes2010geometric, mourant2007optic} shows that most of traffic signs fall in the GFOV equals to $55^{\circ}$ can be recognized accurately in the 60 mph velocity, and most of traffic signs fall in the GFOV equals to $85^{\circ}$  can be recognized accurately in the 30 mph velocities. Linear interpolation was used to calculate the GFOV at different velocities. The value of $E^{sight}$ is changed by the value of the sight line deviation angle under $\eta=6$ as shown in Fig. \ref{fig_sightLineDeviation}. When the angle is less than the GFOV angle, $E^{sight}=1$, else the diver needs to turn his head to see the traffic sign, so $E^{sight}=1$ drastically reduced. If the angle is greater than $\pi/2$, the vehicle has passed the traffic sign, so it equals to 0.
The middle values of SD for different design speeds listed in \cite{yang1999gb} are used as $d_{type}^{sightDistance}$ in the experiments. In the Chinese traffic sign setting standard, SD is the same when the types change. The mounting height of the road side traffic signs and the height of overhead signs are set to the middle value in the state standard \cite{yang1999gb}. VRT for vehicles traveling under 35 mph in less than three-lane environments can be estimated as 8 s; for vehicles traveling over 35 mph in a more complex four- to five-lane environment, at 10 s. Considering that the driving maneuver can be made after the sign location, the $t^{vrt}$ is set to 4 s and 5 s for vehicles speed 30 mph and 40 mph, respectively \cite{bertucci2006sign}. From an experiment of observing the 100 images with actual viewpoint visibility and ideal viewpoint visibility within the VRT time of 20 students, 16 men, and 4 women. We got the threshold $\sigma=0.71$ that all students can recognize all images.

\begin{table}[!t]
	\caption{Descriptions of parameters of datasets}
	\label{table_parametersSet}
	\centering
	\begin{tabular}{|c||c||c||c|}
		\hline
		\textbf{Parameters} & \textbf{ZCABR} & \textbf{LHSR} & \textbf{WPR} \\
		\hline
		$d^{standard}$ & 2 m & 2 m & 2 m \\
		\hline
		$\alpha$ & 0.8 & 0.8 & 0.8 \\
		\hline
		$\beta$ & 0.2 & 0.2 & 0.2 \\
		\hline
		$\lambda$ & 6 & 6 & 6 \\
		\hline
		$\eta$ & 6 & 6 & 6 \\
		\hline
		$V^{field}$ & $90^{\circ}$ & $53.9^{\circ}$ & $72.8^{\circ}$ \\
		\hline
		$V^{fieldI}$ & $85^{\circ}$ & $75^{\circ}$ & $75^{\circ}$ \\
		\hline
		$\gamma$ & 1 & 1 & 1 \\
		\hline
		$\delta$ & 0 & 0 & 0 \\
		\hline
		$\sigma$ & 0.71 & 0.71 & 0.71 \\
		\hline
		$d^{alpha}$ & 0.1 m & 0.1 m & 0.1 m \\
		\hline
		$d^{retina}$ & $17$ mm & $17$ mm & $17$ mm \\
		\hline
		$h^{eye}$ & 1.2 m & 1.2 m & 1.2 m \\
		\hline
		$d^{width}$ & 2 m & 2 m & 2 m \\
		\hline
		$d^{length}$ & 40 & 40 & 40 \\
		\hline
		$w^{shoulder}$ & 0.5 m & 0.5 m & 0.5 m \\
		\hline
		$h$ & 2/4.75 m & 2/4.75 m & 2/4.75 m \\
		\hline
		$\phi$ & $15^{\circ}$ & $15^{\circ}$ & $15^{\circ}$ \\
		\hline
		$\psi$ & $22.5^{\circ}$ & $22.5^{\circ}$ & $22.5^{\circ}$ \\
		\hline
		SD & 45 m & 60 m & 60 m \\
		\hline
		$t^{vrt}$ & 4 s & 5 s & 5 s \\
		\hline
		$v^{85}$ & 25.0 mph & 61.1 mph & 42.2 mph \\
		\hline
	\end{tabular}
\end{table}

\subsection{Calculation Result and Discuss}
\begin{figure*}[!t]
	\centering
	\subfloat[Top view]{\includegraphics[width=1.7in,height=1.7in]{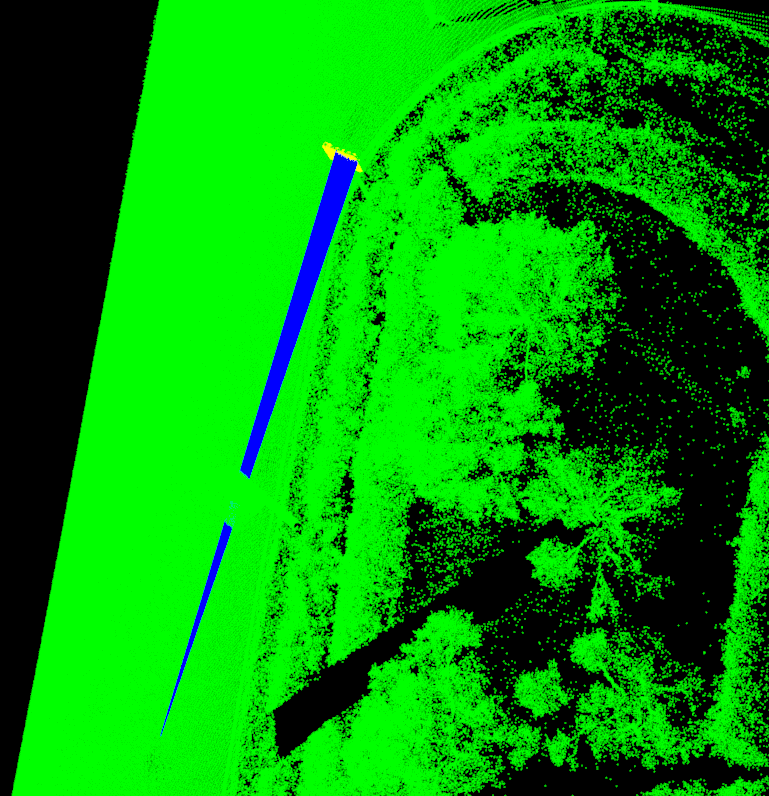}%
		
		\label{figOccludedPointCloudObtainingTopView}}
	\hfil
	\subfloat[Side view]{\includegraphics[width=1.7in,height=1.7in]{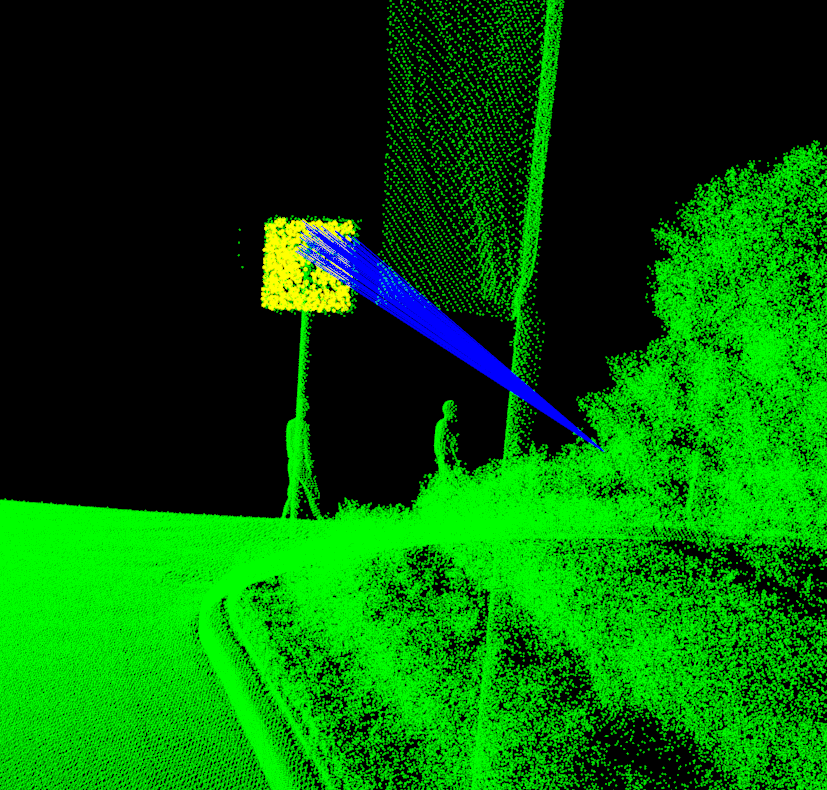}%
		\label{figOccludedPointCloudObtaining}}
	\hfil
	\subfloat[Occlusion area]{\includegraphics[width=1.7in,height=1.7in]{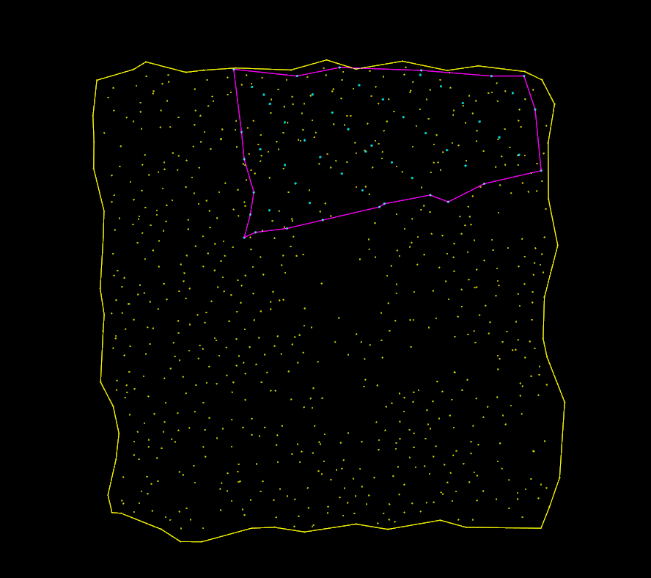}%
		\label{figoccludedArea}}
	\caption{Occluded point cloud obtaining result.}
	\label{figOccludePoints}
\end{figure*}

An example of the calculated results of viewpoint visibility for a traffic sign is illustrated by Fig. \ref{figOccludePoints}. The yellow point cloud is a traffic sign panel. The green point cloud is the traffic sign surrounding objects and road surface point cloud. The common vertex of the blue line cluster is the viewpoint. In Fig. \ref{figoccludedArea}, the closed yellow line is the traffic sign panel point cloud edges, the pink closed line in traffic sign panel is the occlusion part which is occluded by billboard. 

\begin{figure*}[!t]
	\centering
	\includegraphics[width=6.6in]{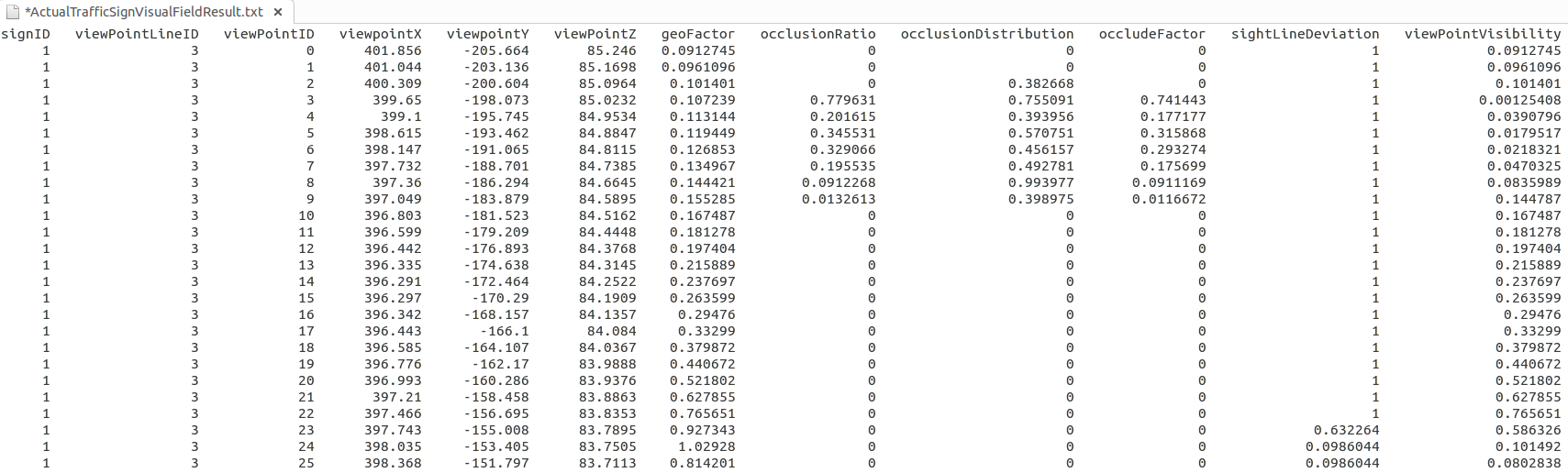}
	\caption{Visibility field results}
	\label{fig_viewpointVisibilityResult}
\end{figure*}

The visibility field results are saved as text, as the Fig. \ref{fig_viewpointVisibilityResult} shows. The viewpoint line is composed by column viewpoints along the road. 

\begin{figure}[!t]
	\centering
	\includegraphics[width=2.5in,height=1.8in]{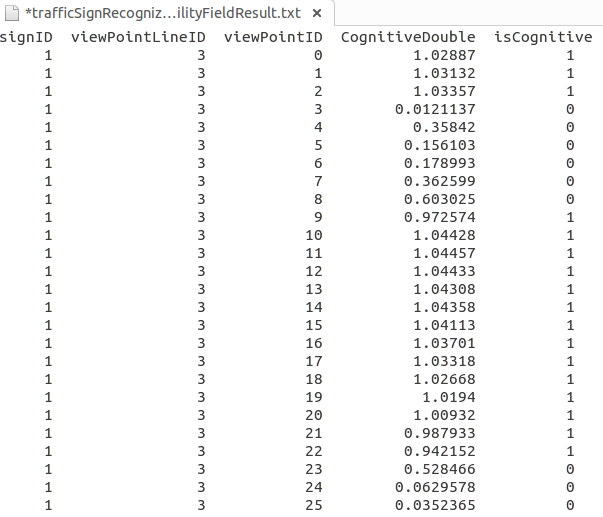}
	\caption{Visual recognizability field results}
	\label{fig_trafficSignRecognizabilityResult}
\end{figure}

The visual recognizability field results are saved in text, as the Fig. \ref{fig_trafficSignRecognizabilityResult} shows. The “\textit{CognitiveDouble}’’ in this figure is the ratio of actual viewpoint visibility to ideal viewpoint visibility. In a normal situation, the “\textit{CognitiveDouble}’’ value is smaller than 1, but sometimes, “\textit{CognitiveDouble}’’ value will be bigger than 1 because the road curve or road surface up and down may lead to the aiming, distance, and sight line from the viewpoint to the traffic sign in the actual traffic environment being better than in the ideal traffic environment. It does not matter for discriminating the recognizability of a viewpoint, if “\textit{CognitiveDouble}’’ value is bigger than $\sigma$, its recognizability equals to 1, else equals to 0.

\begin{figure}[!t]
	\centering
	\includegraphics[width=3.2in,height=1.2in]{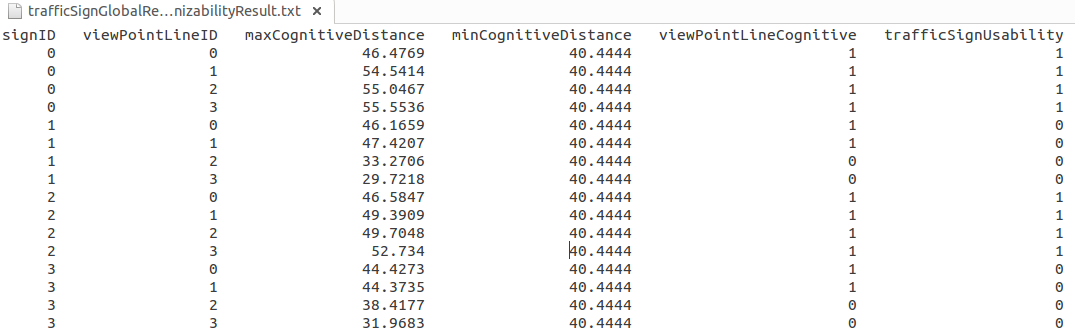}
	\caption{Traffic sign timely visual recognizability result}
	\label{fig_trafficSignGloblalRecognizabilityResult}
\end{figure}

Traffic sign timely visual recognizability results are shown in Fig. \ref{fig_trafficSignGloblalRecognizabilityResult}. The “\textit{maxCognitiveDistance}’’ and “\textit{minCognitiveDistance}’’ in this figure are $d_i^{maxCogLength}$ and $d^{vrd}$ respectively. From this figure, we can see the recognizability of the viewpoint line 3 of traffic sign 1, of which the viewpoint visibilities shown in Fig. \ref{fig_viewpointVisibilityResult} and viewpoint recognizability shown in Fig. \ref{fig_trafficSignRecognizabilityResult} are 0.

\begin{figure}[!t]
	\centering
	\includegraphics[width=3.5 in]{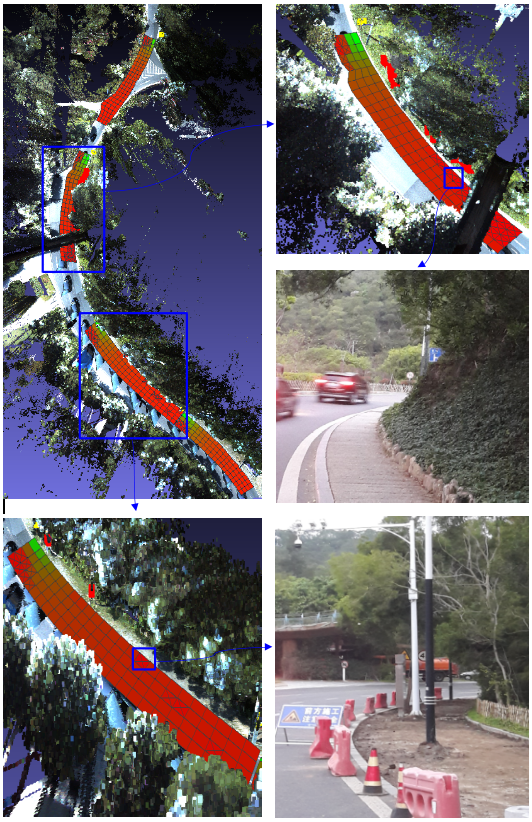}
	\caption{Graphical interface of traffic sign timely visual recognizability results}
	\label{fig_graphicalInterface}
\end{figure}

The Fig. \ref{fig_graphicalInterface} shows a graphical interface to the evaluation of traffic sign timely visual recognizability generated in our algorithm. It is a section of Wenping Road and includes the detected traffic sign (yellow) viewpoint visibility result (mesh plane with color from red to green), viewpoint recognizability result (mesh polygon with color black), occlusion point cloud (red), and some pictures have a same position with corresponding mesh in actual environment. The color of mesh planes change from red to green means that the values of viewpoint visibility change from big to small. The “X’’ symbol in a polygon means that the traffic sign cannot been recognizable in the polygon plane.

The proposed traffic sign timely visual recognizability model was implemented using C++ running on an Intel (R) Core (TM) i5-4460 computer. The computing times in each processing step in every section were recorded in Table \ref{table_timeResult}. As seen from Table \ref{table_timeResult}, among traffic sign detection and classification computing time (STime), road marking detection and classification computing time (MTime) and traffic sign timely visual recognizability evaluation computing time (RTime), traffic sign detection and classification computing time is the least. Road marking detection and classification takes more computing time than traffic sign detection and classification. The computing time of TSTVREM model is related to the number of signs. The time complexity of our proposed method is fast enough to meet the demand of large-scale implementation. This is benefited from the segmentation along the right road outline step, which dramatically reduces the quantity of the point cloud data to be processed. Taking the \textit{sec2} of the WPR dataset with approximately 44 million points and a length of 453.5 m of road, which have 6 traffic signs as an example. The total computing time of our proposed method took 81.339 s to evaluate traffic sign timely visual recognizability from the raw MLS point clouds. Therefore, our method is efficient capable of rapid implementation in a large-scale transportation environment.

\begin{table*}[!t]
	\caption{Computing time of different datasets}
	\label{table_timeResult}
	\centering
	\begin{tabular}{|c||c||c||c||c||c||c||c|}
		\hline
		\textbf{Roads} & \textbf{Sections} & \textbf{Length} & \textbf{PointsNumber} & \textbf{STime} & \textbf{MTime} & \textbf{RTime}  & \textbf{TotalTime} \\
		\hline
		\textbf{ZCABR} 
		& sec1 & 770.06  m & 131681009 & 101.562  s & 1331.77  s & 23.4443 s & 1456.78  s  \\
		\hline
		\multirow{3}{*}{\textbf{LHSR}}
		& sec1 & 734.051 m & 91212406  & 34.9641 s & 462.631 s & 11.0206 s & 508.615 s \\
		\cline{2-8}
		& sec2 & 593.169 m & 69915170  & 24.2266 s & 178.086 s & 3.50449 s & 205.817 s \\
		\cline{2-8}
		& sec3 & 244.527 m & 26623511  & 8.99354 s & 16.3995 s & 1.11341 s & 26.5064 s \\
		\hline
		\multirow{3}{*}{\textbf{WPR}} 
		& sec1 & 453.526 m & 44234114  & 15.2498 s & 0.131961 s & 0.00027 s & 15.3821 s \\
		\cline{2-8}
		& sec2 & 632.114 m & 51933545  & 19.9543 s & 54.6281 s & 6.75658 s & 81.339 s \\
		\cline{2-8}
		& sec3 & 870.5 m   & 34694921  & 28.8413 s & 58.7381 s & 5.59397 s & 93.1734 s\\
		
		\hline
	\end{tabular}
\end{table*}

\section{Conclusion}
This paper presented a traffic sign timely visual recognizability evaluation model for traffic sign inventory and management purpose based on human visual cognition theory and traffic big data using measurable point clouds collected by an MLS system. In the process of building the model, we considered a number of factors, such as traffic sign's size, position, placement, mounting height, panel aiming, depression angle, shape damage, occlusion, actual vehicle speed, sight line deviation, GFOV, VRT, road curve, road surface gradient, and different lanes. The conception of a visibility field is addressed to reflect the traffic sign visibility distribution in 3D space. A visibility evaluation model is presented based human visual cognition theory to compute a traffic sign's visibility for a given viewpoint. Comparing with the concept of a visibility field, we addressed the conception of a visual recognizability field to reflect the visual recognizable distribution in 3D surface and proposed the visual recognition evaluation model to compute a traffic sign's visual recognizability from a given viewpoint. In order to evaluate a traffic sign timely visual recognizability in different lanes, we propose a TSTVREM model by combining the visual recognizability field with the actual maximum continuous cognitive distance and traffic big data. Finally, we constructed an automatic algorithm to implement the TSTVREM model. The algorithm includes: traffic sign and road marking point clouds extraction and classification; traffic sign surrounding point clouds segmentation; viewpoints computation in different lanes; and TSTVREM model realization. In addition, not only for traffic signs, but also for other traffic devices, the timely visual recognizability can be evaluated by our model based on the MLS point cloud. For example, the traffic light. The only different form traffic sign is that it is needed to detect the traffic light from point cloud and think of it as a plane.

Our model is based on three key ideas. First, we evaluate the viewpoint visibility from the angle of human retinal imaging, GFOV changes at different driving speeds and roadway condition, and occlusion degree as other works. Second, we get the actual driving speed from traffic big data to evaluate the timely visual recognizability of the traffic sign. Third, we use the latest equipment obtained the measurable point clouds of roadway environment and use the latest point cloud processing algorithm to enable the implementation of our model.

In the future, other facts may be considered for our model. For example: the light influence caused by solar elevation angle, background influence, and the cognitive burden of traffic density, among others.


%



\section*{Acknowledgment}
We would like to thank the anonymous reviewers for their valuable comments. This work is supported by grants from Natural Science Foundation of China (No.61371144 and No.U1605254) and Natural Science Foundation of Xizang Autonomous Region of China (No.2015ZR-14-16).

\ifCLASSOPTIONcaptionsoff
  \newpage
\fi



\bibliographystyle{IEEEtran}
\bibliography{tits}
%

%
%
%

%

\begin{IEEEbiography}[{\includegraphics[width=1in,height=1.25in,clip,keepaspectratio]{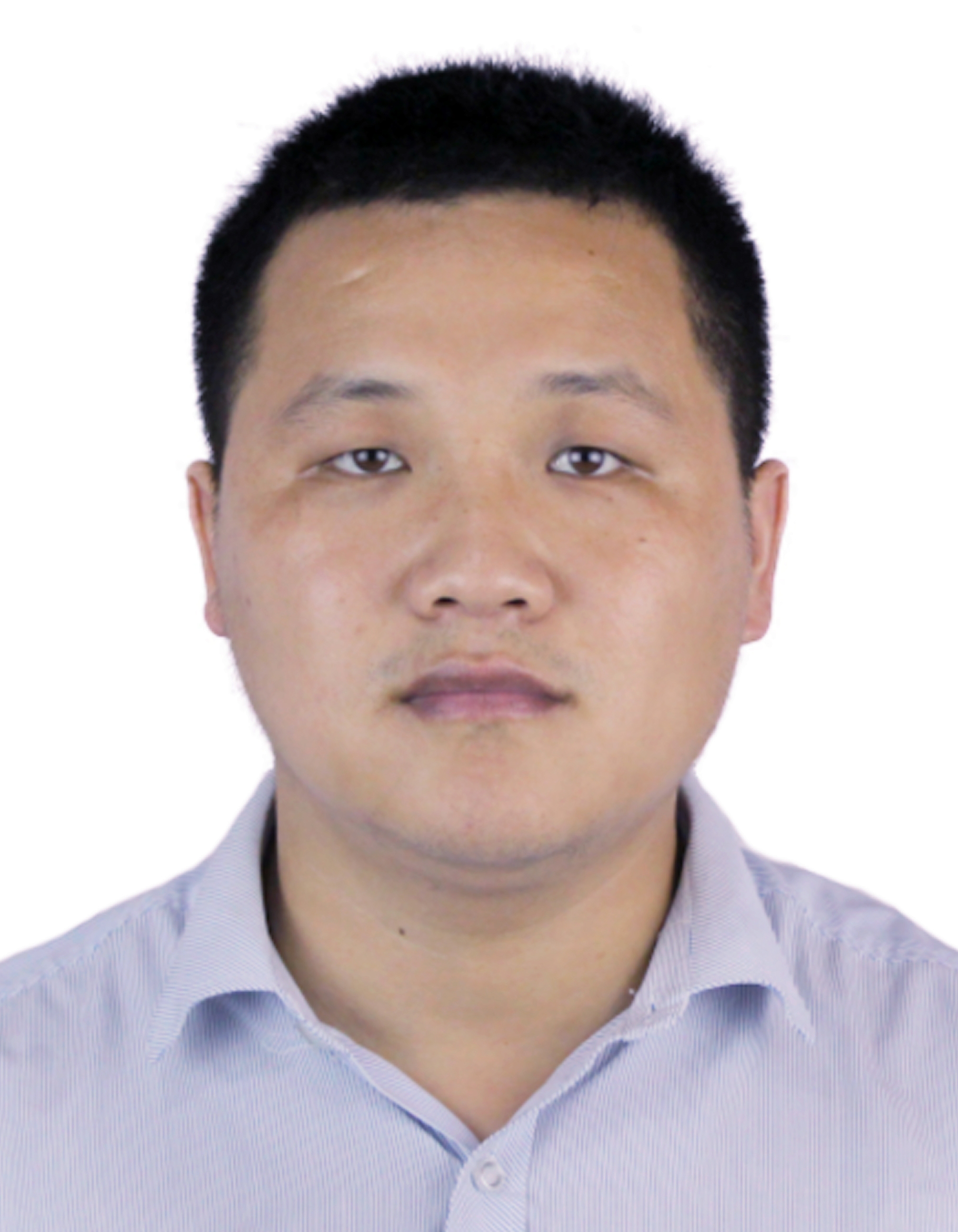}}]{Shanxin Zhang} 
	received his M.E. degree in computer software and theory from Shandong University of Science and Technology, Qingdao, China, in 2010. He is an assistant Professor of Xizang Minzu University and currently a Ph.D. student in computer science and technology with the Fujian Key Laboratory of Sensing and Computing for Smart City in the School of Information Science and Engineering, Xiamen University, China. His current research interests include computer vision, machine learning, deep learning, and mobile LiDAR point clouds data processing.
\end{IEEEbiography}

\begin{IEEEbiography}[{\includegraphics[width=1in,height=1.25in,clip,keepaspectratio]{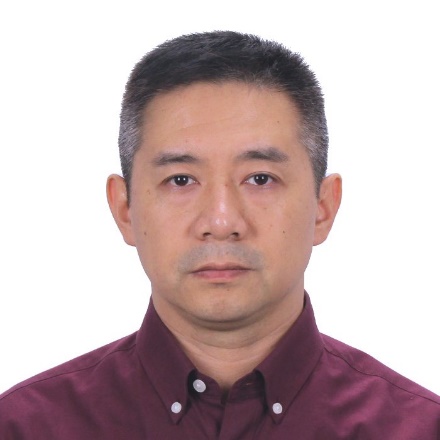}}]{Cheng Wang}
	(M'04-SM'16) received the Ph.D. degree in information communication engineering from the National University of Defense Technology, Changsha, China, in 2002. He is currently Professor and Associate. Dean of the School of Information Science and Engineering, and Executive Director of Fujian Key Laboratory of Sensing and Computing for Smart City, both at Xiamen University, China. His current research interests include remote sensing image processing, mobile LiDAR data analysis, and multi-sensor fusion. He is Chair of ISPRS WG I/6 on Multi-Sensor Integration and Fusion (2016-2020), council member of the China Society of Image and Graphics. He has coauthored more than 150 papers in referred journals and top conferences including IEEE-TGRS, PR, IEEE-TITS, AAAI, and ISPRS-JPRS.
	
\end{IEEEbiography}

\begin{IEEEbiography}[{\includegraphics[width=1in,height=1.25in,clip,keepaspectratio]{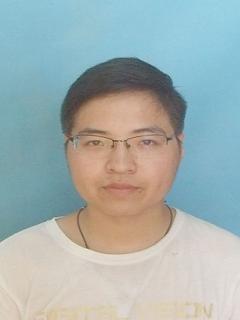}}]{Zhuang Yang} 
	received the M.Sc. degree in computational mathematics from the GuiLin University of Electronic Technology, Guilin, China in 2014. He is currently pursuing a Ph.D. degree in information and communication engineering with the Fujian Key Laboratory of Sensing and Computing for Smart Cities and the Department of Communication Engineering, School of Information Science and Engineering, Xiamen University, Xiamen, China. His current research interests include machine learning, computer vision, matrix analysis, optimization algorithm.
\end{IEEEbiography}

\begin{IEEEbiography}[{\includegraphics[width=1in,height=1.25in,clip,keepaspectratio]{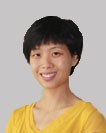}}]{Chenglu Wen}
	received  the Ph.D. Degree in mechanical engineering from China Agricultural University, Beijing, China, in 2009. She is currently an Assistant Professor with Fujian Key Laboratory of Sensing and Computing for Smart City, School of In- formation Science and Engineering, Xiamen University, Xiamen, China. She has coauthored more than 30 research papers published in refereed journals and proceedings. Her current research interests include machine vision, machine learning, and point cloud data processing. She is the Secretary of the ISPRS WG I/3 on Multi-Platform Multi-Sensor System Calibration (2012–2016).
	
\end{IEEEbiography}

\begin{IEEEbiography}[{\includegraphics[width=1in,height=1.25in,clip,keepaspectratio]{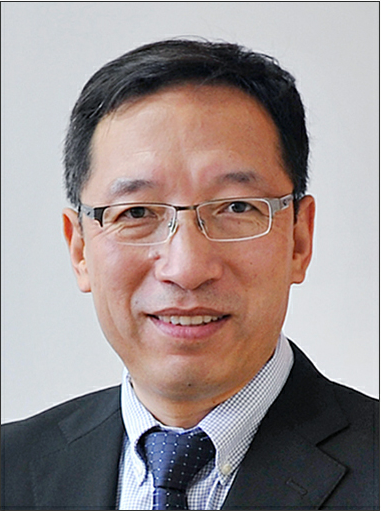}}]{Jonathan Li}
	(M'00-SM'11) received the Ph.D. degree in geomatics	engineering from the University of Cape Town, South Africa. He is currently a Professor and head of the Mobile Sensing and Geodata Science Lab at the Department of Geography and Environmental Management, University of Waterloo, Canada. His current research interests include information extraction from LiDAR point clouds and from earth observation images. He has co-authored more than 350 publications, over 150 of which were published	in refereed journals including IEEE-TGRS, IEEE-TITS, IEEE-GRSL, IEEEJSTARS,	ISPRS-JPRS, IJRS, PE\&RS and RSE. He is Chair of the ISPRS Working Group I/2 on LiDAR for Airborne and Spaceborne Sensing (2016-2020), Chair of the ICA Commission on Sensor-driven Mapping (2015-2019), and Associate Editor of IEEE-TITS and IEEE-JSTARS.
\end{IEEEbiography}

\begin{IEEEbiography}[{\includegraphics[width=1in,height=1.25in,clip,keepaspectratio]{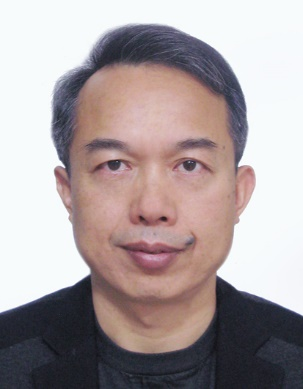}}]{Chenhui Yang}
	 received the Ph.D. degree in Mechanical Engineering from Zhejiang University, Zhejiang, China, in 1995. He is currently a Professor of the School of Information Science and Engineering at Xiamen University, China. He was a visiting scholar in University of Chicago/Argonne National Laboratory in 1999--2000 and in University of Southern California in 2014--2015. His research interests focus on computer vision, computer graphics, data mining, and their applications to other sciences and industries, including transportation, security, and medicine, among others.
\end{IEEEbiography}








\end{document}